\documentclass[letterpaper, 10 pt, conference]{ieeeconf}  

\IEEEoverridecommandlockouts                              

\usepackage{cite}
\usepackage{amsmath,amssymb,amsfonts}
\usepackage{algorithmic}
\usepackage{graphicx}
\usepackage{textcomp}
\usepackage{multirow}
\usepackage{makecell}
\usepackage{booktabs}
\usepackage{tikz}
\usepackage{pgfplots}
\usepackage[figuresright]{rotating}
\usepackage{extarrows}
\usepackage{xcolor}
\usepackage{placeins}   
\usepackage{float}
\usepackage{siunitx}
\title{\LARGE \bf
	QuietWalk: Physics-Informed Reinforcement Learning for Ground Reaction Force-Aware Humanoid Locomotion Under Diverse Footwear
}

\author{Hanze Hu$^{1,2,3}$, Luying Feng$^{4}$, Silu Chen$^{1,2,3,*}$, Tianjiang Zheng$^{1,2,3}$, Dexin Jiang$^{1,2,3}$,\\
	Wei Chen$^{4}$, Chi Zhang$^{1,2,3}$, Guilin Yang$^{1,2,3}$, Yaochu Jin$^{4,*}$%
	\thanks{$^{1}$Ningbo Institute of Materials Technology and Engineering, Chinese Academy of Sciences, Ningbo 315201, China.}%
	\thanks{$^{2}$University of Chinese Academy of Sciences, Beijing 100049, China.}%
	\thanks{$^{3}$Zhejiang Key Laboratory of Precision Actuation and Intelligent Robotics.}%
	\thanks{$^{4}$School of Engineering, Westlake University, Hangzhou, China.}%
	\thanks{$^{*}$Silu Chen and Yaochu Jin are the corresponding authors. E-mail: chensilu@nimte.ac.cn; jinyaochu@westlake.edu.cn}%
}

\begin{document}

	\maketitle
	\thispagestyle{empty}
	\pagestyle{empty}

	\begin{abstract}
		Humanoid robots operating in human-centered environments (e.g., homes, hospitals, and offices) must mitigate foot--ground impact transients, as impact-induced vibration and noise degrade user experience and repeated impacts accelerate hardware wear. However, existing low-noise locomotion training often relies on kinematic proxy objectives or fragile force sensors, and footwear-induced changes in contact dynamics introduce distribution shifts that hinder policy generalization.
		We present \textbf{QuietWalk}, a physics-informed reinforcement learning framework for ground-reaction-force-aware humanoid locomotion under diverse footwear conditions. QuietWalk employs an inverse-dynamics-constrained physics-informed neural network (PINN) to estimate per-foot vertical ground reaction forces (GRFs) from proprioceptive signals, and integrates the frozen predictor into the RL training loop to penalize predicted impact forces without requiring force sensors at deployment.
		On a held-out real-robot dataset, enforcing inverse-dynamics consistency reduces vertical GRF prediction errors by 82--86\% compared with a purely supervised predictor and improves the coefficient of determination from 0.39/0.67 to 0.99/0.99 for the left/right feet. On hardware at 1.2\,m/s (barefoot; averaged over four floor materials), QuietWalk reduces mean A-weighted noise level by 7.17\,dB and peak noise level by 4.98\,dB under a consistent recording setup. Cross-footwear experiments (barefoot, skate shoes, athletic sneakers, and high heels) across multiple surfaces further demonstrate robust adaptation to footwear-induced contact variations.
	\end{abstract}

	\section{INTRODUCTION}
	With the rapid development of humanoid robotics, application scenarios are expanding from industrial settings to human living spaces such as homes, hospitals, and offices. In these human-centered environments, robots must not only achieve basic locomotion, but also meet stricter acoustic and safety requirements \cite{zhang2021exploring}. During walking, the ground reaction force (GRF) generated by foot--ground contact is a primary source of structural vibration and acoustic noise \cite{cao2025minimizing}. Excessive impact forces not only produce unpleasant noise that disrupts daily life, but also accelerate wear of robot joints and may even damage the floor due to repeated impacts \cite{ahn2023learning}. Therefore, achieving low-impact, low-noise ``quiet walking'' is a key prerequisite for humanoids to integrate into human society. Moreover, humans exhibit remarkable locomotion adaptability: they can maintain stable gait while wearing footwear with different properties \cite{chen2020influences}. Enabling robots with similar cross-footwear generalization, so that they can adapt to changes in contact surfaces and dynamics induced by different shoes, is an important direction toward improved general-purpose capability.
	
	\begin{figure}[!t]
		\centering
		\includegraphics[width=0.9\columnwidth]{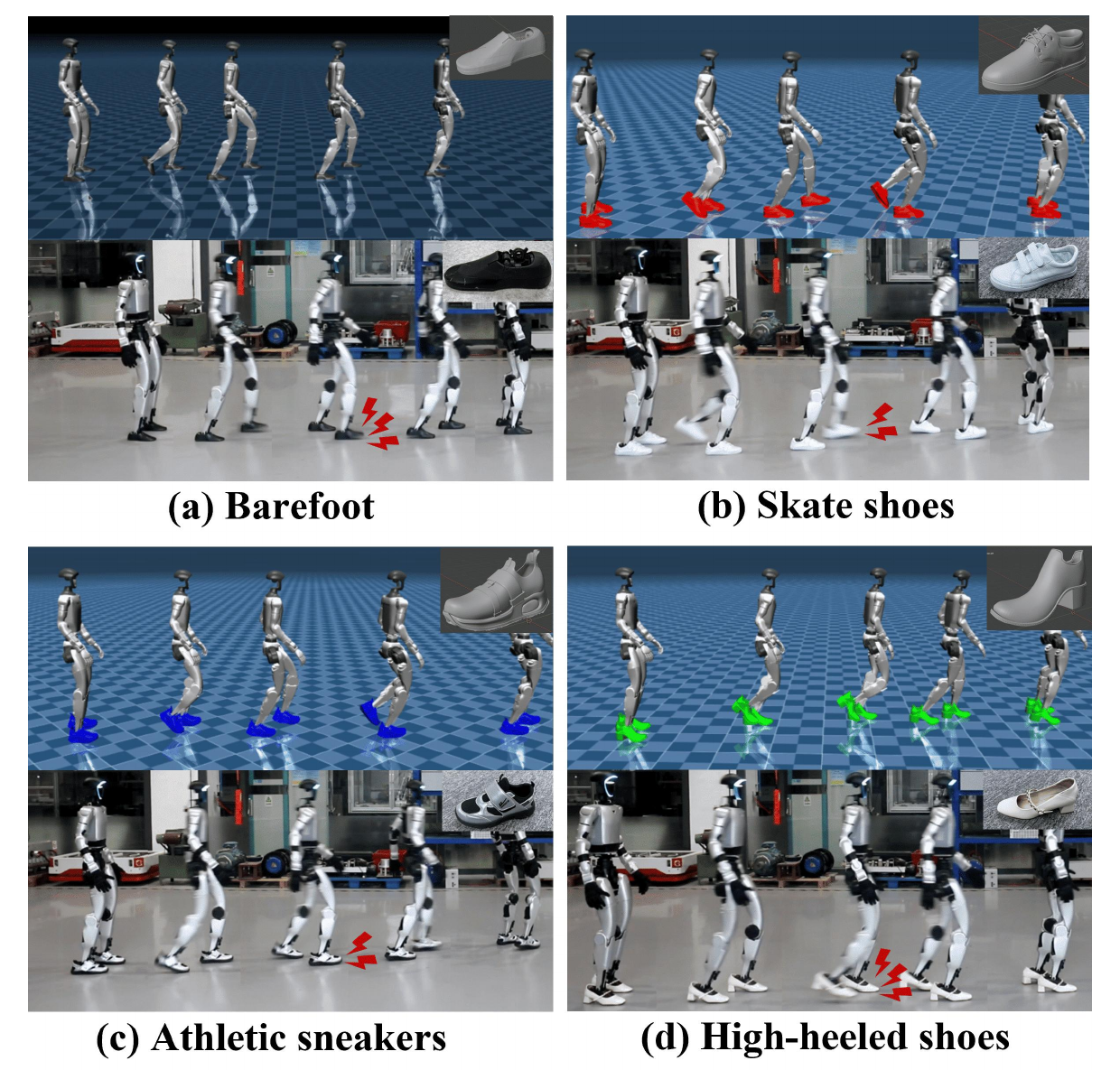}
		\caption{Humanoid robot validation in simulation and on hardware under barefoot, skate shoes, athletic sneakers, and high heels.}
		\label{fig:overview}
		\vspace{-0.8em} 
	\end{figure}
	
	\begin{figure*}[!t]
		\centering
		\includegraphics[width=0.9\textwidth]{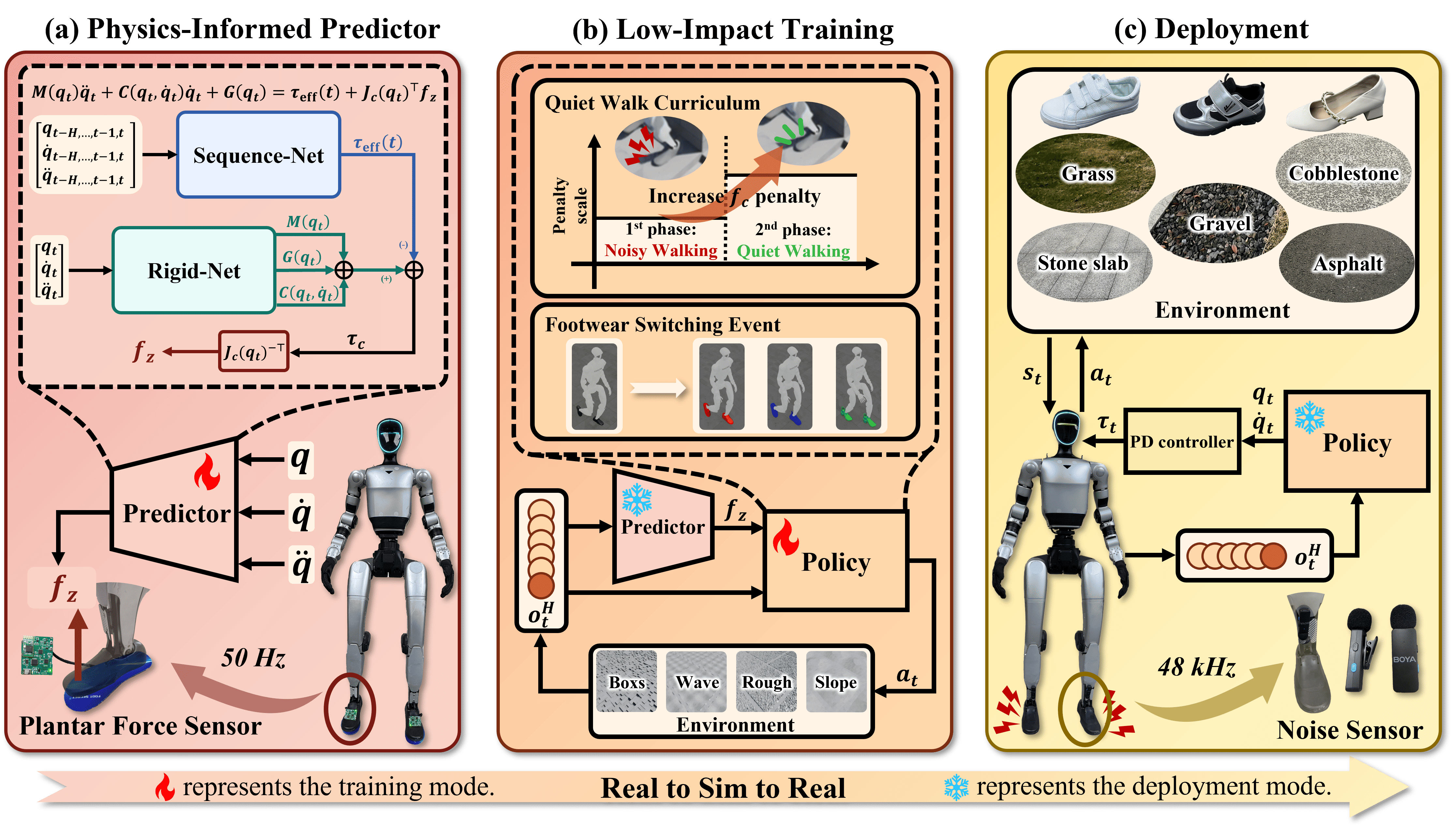}
		\caption{Overview of the proposed physics-informed reinforcement learning framework with cross-footwear deployment.}
		\label{fig:method}
		\vspace{-0.8em} 
	\end{figure*}

	Although reinforcement-learning-based controllers have made significant progress in robust legged locomotion in recent years, and have demonstrated strong adaptability under complex terrains and disturbances \cite{margolis2024rapid,li2025reinforcement}, research targeting the ``quiet/low-impact'' objective required by human-centered scenarios such as homes remains limited. Prior work often relates footstep noise to kinematic quantities such as foot contact velocity, and indirectly reduces impact and noise by penalizing contact velocity or velocity changes in the reward function \cite{watanabe2025learning}. Meanwhile, because GRF is difficult to measure directly, researchers have proposed physics-based inverse-dynamics estimation and deep-learning-based force prediction methods to improve the usability and consistency of force estimation \cite{forner2006inverse,karatsidis2016estimation}. However, these approaches still face three key bottlenecks. First, traditional force-sensor signals are noisy and fragile \cite{hashlamon2013ground}; using them directly in RL rewards can cause training instability or even failure to converge. Second, relying only on contact-velocity penalties often sacrifices gait agility and stability \cite{zhang2025quietpaw}, making it difficult to balance ``quiet'' walking with ``fast/robust'' locomotion. Third, footwear changes (especially contact-dynamics changes induced by high heels) introduce significant distribution shift \cite{hyun2016effect}, which limits policy generalization, and systematic cross-footwear robustness validation is still lacking.
	
	This paper aims to achieve low-impact humanoid walking with explicit cross-footwear generalization via a coupled framework of ``physics-informed GRF estimation + reinforcement learning''. Specifically, we propose a PINN-based GRF predictor that incorporates inverse-dynamics constraints into the network, enabling accurate and low-noise contact-force estimation without force sensors and thus providing more reliable force signals for learning. Furthermore, we build a low-impact RL framework that embeds the GRF predictor into the training loop as a reward critic. It directly penalizes high-impact transients while preserving gait stability and locomotion performance. Finally, we propose a cross-footwear robustness test protocol to systematically evaluate generalization across footwear conditions (from barefoot to high heels) and to reveal how changes in contact properties affect learned gaits. These contributions provide a feasible path toward quieter, gentler, and more durable indoor humanoid locomotion.
	
	In summary, the main contributions of this paper are as follows:
	
	\begin{itemize}\relax
		\item \textbf{Inverse-Dynamics-Constrained GRF Predictor:}
		We develop a physics-informed neural network that enforces inverse-dynamics consistency for vertical GRF estimation using only proprioceptive inputs, reducing reliance on noisy force sensors and improving physical consistency compared with purely data-driven predictors.
	\end{itemize}
	
	\begin{itemize}\relax
		\item \textbf{Physics-Informed Impact-Aware Reinforcement Learning:}
		We integrate the frozen, inverse-dynamics-consistent GRF predictor into the RL loop as a physics-consistent reward critic, enabling impact-aware policy optimization with consistent force feedback between training and deployment, without external force sensing.
	\end{itemize}
	
	\begin{itemize}\relax
		\item \textbf{Cross-Footwear Validation:}
		This work introduces footwear-induced contact variation as a new robustness dimension for humanoid locomotion control.
	\end{itemize}

	\section{RELATED WORKS}
	\subsection{Physics-Informed Learning for Dynamics Estimation}
	
	Data-driven methods have shown strong potential for modeling complex robot dynamics that are difficult to describe analytically. However, purely data-driven models often generalize poorly outside the training distribution. To address this issue, physics-informed neural networks (PINNs) have emerged as a paradigm shift. They embed physical laws directly into the network architecture or the loss function \cite{raissi2019physics}. In robotics, this concept has been extended to Lagrangian neural networks (LNNs) to model non-conservative forces such as friction and damping \cite{djeumou2022neural,liu2024physics}. In legged locomotion, accurate estimation of the ground reaction force (GRF) is crucial for stability. Some works rely on deep learning to map inertial measurement unit (IMU) data to GRF \cite{hossain2023estimation}, but these methods lack physical consistency. Recent advances, such as Mysteric-Net \cite{yeo2025mysteric}, combine LNNs with temporal convolutional networks (TCNs) to capture hysteretic friction and rigid-body dynamics for force estimation. In addition, PINNs have been used to solve inverse problems in contact mechanics by enforcing inequality constraints (Karush--Kuhn--Tucker (KKT) conditions) \cite{sahin2024solving}. Despite these successes, applying PINN-based inverse dynamics to real-time GRF estimation within a reinforcement learning (RL) closed loop remains underexplored, especially for sensorless force feedback in humanoid robots.
	
	\subsection{Control Strategies for Acoustic Stealth and Impact Reduction}
	
	Reducing ground impact is essential not only for hardware protection, but also for acoustic stealth in human--robot shared environments. Traditional model-based approaches mainly reduce landing impulse via impedance control \cite{guadarrama2022preemptive,zhu2024impedance} or by optimizing gait parameters \cite{ahn2023learning}. In RL, ``quiet walking'' is typically achieved indirectly by penalizing high foot-end velocity or acceleration in the reward function. For example, Sony Group Corporation's Aibo \cite{watanabe2025learning} and Disney Research's Olaf \cite{muller2025olaf} successfully reduced walking noise by penalizing foot contact velocity, but they also reported a trade-off between quietness and robustness. Similarly, QuietPaw \cite{zhang2025quietpaw} proposed a noise-constrained policy that can dynamically adjust the ``quietness level'' based on environmental requirements. Other work has attempted to explicitly integrate acoustic metrics into the control loop to minimize noise generation \cite{cao2025minimizing}. However, these methods often rely on kinematic proxy variables rather than explicit force control, or require costly external microphone setups. A framework is still missing that leverages accurate, physics-based force estimation to directly penalize impact forces. Such a framework would enable a precise balance between locomotion performance and acoustic stealth without relying on fragile physical force sensors.
	
	\subsection{Robustness to Morphological and Environmental Variations}
	
	Robust locomotion control must adapt not only to changes in external terrain, but also to changes in the robot's own morphological properties. Extensive research has focused on terrain adaptation. It uses proprioceptive history \cite{kumar2022adapting,nahrendra2023dreamwaq,radosavovic2024real} or fuses proprioception with exteroceptive sensing \cite{miki2022learning,agarwal2023legged} to achieve locomotion in complex environments. However, a less explored dimension is adaptation to end-effector variations, which is analogous to ``changing shoes'' for robots. Mechanical properties of the feet, such as stiffness and damping, can significantly alter energy consumption and stability margins during walking \cite{schumann2019effects}. In biomechanics, it is well known that humans substantially adjust postural control strategies when wearing different footwear to maintain stability \cite{chen2020influences}. Although adaptive feet such as SoftFoot \cite{piazza2024analytical} and footwear technologies with adjustable compliance \cite{price2025adjustable} have been developed to study these effects, current robot control policies rarely account explicitly for the drastic changes in contact dynamics induced by different footwear. This work fills this gap by proposing a framework that leverages physics-informed force estimation. It enables general-purpose locomotion control that is robust to both environmental changes and footwear-induced dynamics variations.

	\section{METHODS}
	To enable deployable low-impact walking control under different footwear contact conditions, the proposed method consists of three components. First, we develop a PINN with robot inverse-dynamics constraints that estimates the bilateral normal GRF using only proprioception. Second, we provide a geometry and inertia parameterization pipeline for footwear assets, so that footwear differences can be reproduced consistently across simulators. Third, we use the GRF predictor for physics-informed reward shaping to train a low-impact reinforcement learning gait policy.
	
	\subsection{Inverse-Dynamics-Constrained PINN for GRF Prediction}
	
	At each control step, we use the proprioceptive history $[q;\dot{q};\ddot{q}]$ as input, where the joint dimension is $n=29$ and
	$q,\dot{q},\ddot{q}\in\mathbb{R}^{n}$. The concatenated feature is thus $[q;\dot{q};\ddot{q}]\in\mathbb{R}^{3n}$.
	To capture short-term transients around contact switches, we stack the most recent 6 frames (50\,Hz, corresponding to 0.12\,s). The network outputs the bilateral normal GRF
	$f_z=[f_z^{(L)},f_z^{(R)}]^T\in\mathbb{R}^2$. In the training set, the bilateral normal GRF is collected at 50\,Hz using force sensors mounted under the robot feet.
	
	\begin{figure}[H]
		\centering
		\includegraphics[width=1.0\columnwidth]{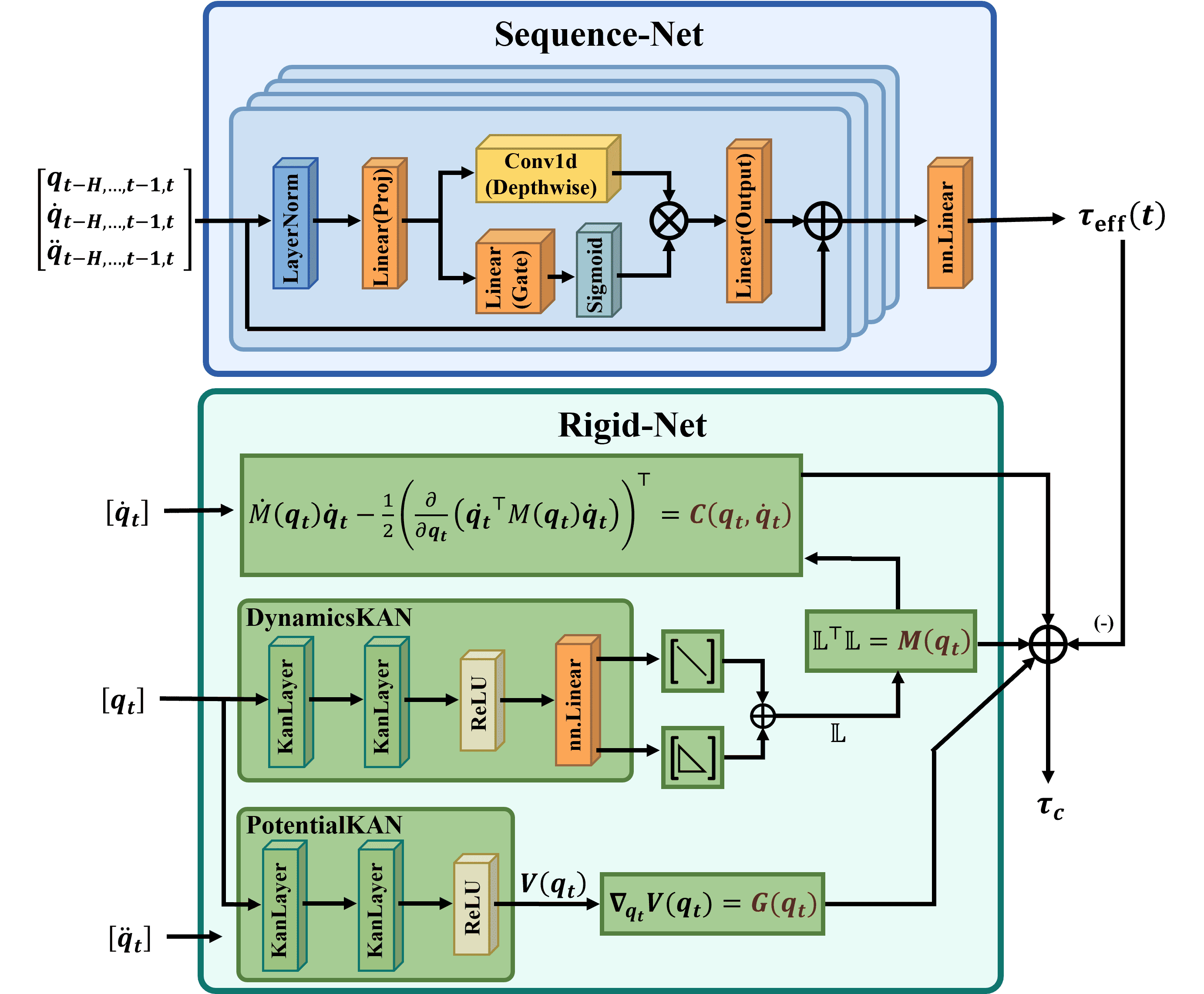}
		\caption{Structure of PINN-Based robot inverse dynamics.}
		\label{fig:network}
		\vspace{-0.8em} 
	\end{figure}	
	
	The network architecture is shown in Fig.~\ref{fig:network}. To avoid fully black-box fitting, we explicitly constrain the dynamic structure and learn only the necessary mappings. DynamicsKAN predicts the inertia matrix
	$M(q)\in\mathbb{R}^{n\times n}$ from $q$, and enforces symmetry and positive definiteness via a factorized form \cite{lutter2019deep}. PotentialKAN learns the scalar potential
	$V(q)$ and sets $G(q)=\nabla_q V(q)$, which preserves the conservative structure. The Coriolis/centrifugal generalized force is obtained from the learned $M(q)$ via additional matrix operations \cite{park1995lie} to compute $C(q,\dot{q})$. In parallel, Sequence-Net estimates the effective generalized actuation
	$\tau_{\mathrm{eff}}\in\mathbb{R}^{n}$, which absorbs joint dynamic friction and unmodeled non-contact effects. Here, KANLayer is the basic mapping layer of Kolmogorov--Arnold networks (KAN) \cite{kilani2024kolmogorov}. It parameterizes each input feature nonlinearly using learnable B-splines on a predefined grid, and realizes a differentiable mapping from batched input features to batched output features through spline coefficients.
	
	Under normal contact conditions, the system satisfies in joint space
	\begin{align}\label{eq:dyn_ieee}
		M(q)\ddot{q}+C(q,\dot{q})\dot{q}+G(q)=\tau_{\mathrm{eff}}+J_n^T f
	\end{align}
	where $J_n\in\mathbb{R}^{2\times n}$ is the bilateral normal contact Jacobian, and $f=[f^{(L)},f^{(R)}]^T$ is the normal force to be estimated. This defines the contact-induced generalized force $\tau_c$. To obtain numerically stable contact-force reconstruction when the contact Jacobian is ill-conditioned (e.g., single-foot contact or near-singular configurations), we solve for the contact force $f_{\mathrm{raw}}$ using a damped least-squares (DLS) regularized pseudoinverse \cite{chiaverini1994review}.
	\begin{align}\label{eq:tau_c_ieee}
		&\tau_c = M(q)\ddot{q} + C(q,\dot{q})\dot{q} + G(q) - \tau_{\mathrm{eff}} \\
		&f_{\mathrm{raw}} = (J_n J_n^{T} + \lambda^2 I)^{-1} J_n \tau_c
	\end{align}
	where $\lambda > 0$ is a damping coefficient that improves numerical stability under ill-conditioned or near-singular contact configurations.
	
	To enforce unilateral contact constraints, the predicted normal force is projected onto the non-negative orthant as $f_z=\max(f_{\mathrm{raw}},0)$, which ensures physically consistent non-negative normal contact forces.
	
	During training, we minimize a weighted loss that jointly optimizes supervised GRF regression and inverse-dynamics residuals, with optional swing-phase output and temporal smoothing:
	
	\begin{equation}
		\begin{split}
			\mathcal{L} &=
			\lambda_{\mathrm{grf}}\mathcal{L}_{\mathrm{grf}}
			+\lambda_{\mathrm{dyn}}\mathcal{L}_{\mathrm{dyn}}
			+\lambda_{\mathrm{swing}}\mathcal{L}_{\mathrm{swing}}
			+\lambda_{\mathrm{smooth}}\mathcal{L}_{\mathrm{smooth}}
		\end{split}
	\end{equation}
	where
	\begin{align*}
		&\mathcal{L}_{\mathrm{grf}} = \bigl\|\tilde f_z^{\mathrm{pred}}-\tilde f_z^{\mathrm{gt}}\bigr\|^2 ,\\
		&\mathcal{L}_{\mathrm{dyn}} = \bigl\|\tau_c - J_n^\top f_z^{\mathrm{pred}}\bigr\|^2 ,\\
		&\mathcal{L}_{\mathrm{swing}} = \bigl\|(1-c)\odot(\tilde f_z^{\mathrm{pred}}-\tilde 0)\bigr\|^2 ,\\
		&\mathcal{L}_{\mathrm{smooth}} =
		\frac{1}{H-1}\sum_{t=2}^{H}\bigl\|f_z^{\mathrm{pred}}(t)-f_z^{\mathrm{pred}}(t-1)\bigr\|^2 .
	\end{align*}
	
	The predicted normal GRF is $f_z^{\mathrm{pred}}=\max(f_{\mathrm{raw}},0)\in\mathbb{R}^2$, and the ground-truth normal GRF is denoted as $f_z^{\mathrm{gt}}\in\mathbb{R}^2$, both in N.
	To improve optimization stability, we compute the supervised regression term in the normalized force space:
	$\tilde f_z=(f_z-\mu_f)/\sigma_f$, where $(\mu_f,\sigma_f)$ are the mean and standard deviation of the normal GRF estimated from the training set.
	Thus,
	$\tilde f_z^{\mathrm{pred}}=(f_z^{\mathrm{pred}}-\mu_f)/\sigma_f$,
	$\tilde f_z^{\mathrm{gt}}=(f_z^{\mathrm{gt}}-\mu_f)/\sigma_f$.
	Further, we define a binary contact mask for each foot as $c\in\{0,1\}^2$. It is obtained by thresholding
	$f_z^{\mathrm{gt}}$, and we use $(1-c)$ as the swing-phase mask.
	The operator $\odot$ denotes element-wise masking.
	In addition, $\tilde 0=(0-\mu_f)/\sigma_f$ is the normalized value corresponding to zero force.
	For sequence inputs with length $H=6$, $\mathcal{L}_{\mathrm{smooth}}$ penalizes temporal variation; for single-step inputs, we set
	$\mathcal{L}_{\mathrm{smooth}}=0$.
	
	\subsection{Footwear Asset Modeling and Parameterization}
	
	We restrict footwear effects to changes in foot-end geometry and inertia. Based on the Unitree G1 foot model, we construct three footwear assets (Skate shoes, Athletic sneakers, High-heeled shoes), and merge them with the foot link to form a unified ``foot+shoe'' geometry.
	In Blender, we align the shoe meshes such that, in upright standing, the outsole lies flat on the ground. We then merge the shoe and foot meshes using Boolean union to eliminate gaps and interpenetration, and adjust key geometry according to shoe type (outsole thickness, toe curvature, heel height). We compensate for the effective leg-length change introduced by outsole thickness to avoid contact penetration or floating.
	We export the merged geometry as STL and integrate it into URDF, MJCF, and USD formats, while synchronizing both visual and collision geometries. Under a unified density assumption, we recompute the mass and inertia of the merged geometry to ensure consistent inertia distribution across simulators. This maximally isolates footwear differences and reduces implementation artifacts.
	
	\subsection{Physics-Informed Low-Impact RL for Quiet Locomotion}
	
	We train the low-impact walking policy in Isaac Sim and use a pretrained GRF estimator to provide a deployable impact-feedback signal for reward shaping. The policy runs at 50\,Hz, takes 6 frames of proprioceptive history as input, and outputs joint position increments $\Delta q\in\mathbb{R}^{n}$, which are tracked by a PD controller:
	\begin{align}
		&q_{\mathrm{des}} = q_{\mathrm{cur}} + \Delta q \\
		&\tau_{\mathrm{cmd}} = K_p\!\left(q_{\mathrm{des}} - q_{\mathrm{cur}}\right) - K_d \dot{q}_{\mathrm{cur}}
	\end{align}
	where $q_{\mathrm{cur}},\dot{q}_{\mathrm{cur}}\in\mathbb{R}^{n}$ are the measured joint positions and joint velocities,
	$q_{\mathrm{des}}\in\mathbb{R}^{n}$ is the target joint position, and $\tau_{\mathrm{cmd}}\in\mathbb{R}^{n}$ is the PD torque command.
	$K_p$ and $K_d$ are fixed stiffness and damping matrices.
	
	The total RL reward is:
	\begin{align}
		r_t=r_{\mathrm{task}}+r_{\mathrm{bonus}}+r_{\mathrm{impact}}
	\end{align}
	where $r_{\mathrm{task}}$ includes standard terms such as velocity tracking and stability, $r_{\mathrm{bonus}}$ includes auxiliary regularizers (e.g., joint energy and motion smoothness), and $r_{\mathrm{impact}}$ is the quiet-locomotion reward based on the physics predictor.
	
	To explicitly penalize impact landing, we use the per-foot normal GRF predicted by the frozen estimator and define $r_{\mathrm{impact}}$ as:
	\begin{align}
		r_{\mathrm{impact}}=-\alpha\big((f_z^{(L)})^2+(f_z^{(R)})^2\big)
	\end{align}
	
	To avoid degenerate solutions and improve generalization, we adopt a staged curriculum and multi-footwear switching events. We first learn stable walking with a small $\alpha$, and then gradually increase $\alpha$ to reduce impact while maintaining task performance. We also progressively increase terrain randomization difficulty from flat ground, and introduce skate shoes/athletic sneakers/high heels starting from barefoot. In the final stage, we sample all footwear types jointly to obtain a single policy that adapts to footwear-induced contact variations. The simulation runs at 500\,Hz and the control loop runs at 50\,Hz; the policy is trained with a PPO-style actor--critic. The GRF predictor is frozen during RL training and is used only for reward computation. This ensures information consistency between training and potential real-world deployment (without external force-sensing devices).

	\section{EXPERIMENTS}
	This study validates the proposed method through a series of real-world experiments. We focus on the following claims: (1) evaluate the accuracy and stability of the GRF predictor, and verify whether it can provide accurate force feedback without external force sensors; (2) verify whether the physics-informed reinforcement learning control policy can achieve low-noise walking without sacrificing gait stability and maneuverability; (3) examine robustness to cross-footwear variations, i.e., whether the proposed method can maintain good gait performance when the robot transitions from barefoot to multiple footwear types (including the extreme case of high heels).
	
	\subsection{Experimental Setup}
	1) Platform: All walking policies are trained in Isaac Sim. The physics simulation step runs at 500\,Hz and the policy update rate is 50\,Hz.
	The policy input is 6 frames of proprioceptive history (0.12\,s). The output is joint position increments
	$\Delta q\in\mathbb{R}^{n}$ (with $n=29$), which are tracked by a fixed PD controller (see Sec.~III-C).
	
	2) Data collection: To train the GRF predictor, we collect joint states $[q;\dot{q};\ddot{q}]\in\mathbb{R}^{3n}$ together with the corresponding bilateral normal ground reaction forces $f_z=[f_z^{(L)},f_z^{(R)}]^T\in\mathbb{R}^2$ during real-robot walking trials. The joint states are obtained from onboard proprioceptive sensing, while $f_z$ is measured using plantar force-sensor insoles mounted under both feet. Each insole adopts a monolithic embedded design comprising eight independent sensing units for force acquisition. Both proprioceptive signals and force measurements are synchronized and sampled at 50\,Hz (see module (a) in Fig.~\ref{fig:method}).
	The collected dataset is split into disjoint training and validation sets obtained from separate recording sessions to reduce temporal and distribution leakage. The test set consists of held-out motion sequences not used during model development. The predictor is trained for 200 epochs using the Adam optimizer with a learning rate of $1\times10^{-3}$ and a sequence length of $H=6$. The model with the lowest validation loss is selected for deployment. All models are trained on a single NVIDIA RTX 5090 GPU.
	
	3) Footwear conditions: Barefoot and three footwear assets (Skate shoes, Athletic sneakers, High-heeled shoes) are constructed using a unified ``foot+shoe'' modeling pipeline (see Sec.~III-B). To evaluate robustness, we test both in simulation and on hardware (see Fig.~\ref{fig:overview}).
	
	4) Acoustic measurement: During walking, we record audio at a fixed location using cinema-grade wireless microphones with a sampling rate of 48\,kHz. We focus on foot--ground contact noise. Therefore, the microphones are placed at approximately 15\,cm above the ground, with one microphone mounted on each of the left and right legs (see module (c) in Fig.~\ref{fig:method}). To accurately assess perceived noise levels, our analysis focuses on the human-audible band from 20\,Hz to 20\,kHz \cite{hoeppner2012ncbi}.

	\subsection{Evaluation Metrics}
	
	\paragraph{GRF metrics.}
	Since this paper predicts only the vertical component of the ground reaction force $f_z$,
	both the prediction and the ground truth are scalars.
	Let $\hat{y}_i$ denote the predicted vertical GRF at time step $i$,
	$y_i$ denote the corresponding measured value,
	and $N$ denote the number of test samples.
	
	We use the following three metrics to evaluate prediction accuracy:
	
	\textbf{1) Root Mean Squared Error (RMSE)}
	\begin{equation}
		\mathrm{RMSE}
		=
		\sqrt{
			\frac{1}{N}
			\sum_{i=1}^{N}
			(\hat{y}_i - y_i)^2
		}.
	\end{equation}
	
	\textbf{2) Mean Absolute Error (MAE)}
	\begin{equation}
		\mathrm{MAE}
		=
		\frac{1}{N}
		\sum_{i=1}^{N}
		\left|
		\hat{y}_i - y_i
		\right|.
	\end{equation}
	
	\textbf{3) Coefficient of Determination ($R^2$)}
	\begin{equation}
		R^2
		=
		1
		-
		\frac{
			\sum_{i=1}^{N}
			(\hat{y}_i - y_i)^2
		}{
			\sum_{i=1}^{N}
			(y_i - \bar{y})^2
		},
		\qquad
		\bar{y}=\frac{1}{N}\sum_{i=1}^{N} y_i.
	\end{equation}
	
	RMSE and MAE are error metrics. Smaller values indicate smaller deviation between predictions and measurements and thus higher accuracy.
	Compared with MAE, RMSE is more sensitive to large errors. It therefore highlights prediction errors during impact phases.
	The coefficient of determination $R^2$ measures how much variance in the measurements is explained by the model,
	with a range of $(-\infty, 1]$.
	$R^2=1$ indicates a perfect fit.
	$R^2=0$ means the model is equivalent to a mean predictor.
	$R^2<0$ indicates performance worse than the mean baseline.
	Therefore, in all result tables, we treat smaller RMSE/MAE and $R^2$ closer to 1 as improved performance.

	\paragraph{Acoustic metrics.}
	To evaluate acoustic performance during walking, we use A-weighted sound pressure level (A-weighted SPL) \cite{fletcher1933loudness}, measured in dBA:
	
	\begin{equation}
		\mathrm{SPL}(\mathrm{dBA})
		=
		20\log_{10}\left(\frac{p}{p_0}\right),
		\quad
		p_0 = 20\,\mu\mathrm{Pa},
	\end{equation}
	
	where $p$ is the root-mean-square (RMS) sound pressure, and $p_0$ is the reference sound pressure corresponding to the human hearing threshold.
	
	Based on the SPL time series over the walking segment, we report:
	
	\begin{itemize}
		\item \textbf{MNL (Mean Noise Level)}: the average SPL over the walking segment, reflecting overall noise exposure.
		\item \textbf{PNL (Peak Noise Level)}: the maximum instantaneous SPL over the walking segment, characterizing peak noise induced by foot--ground impacts.
	\end{itemize}
	
	MNL captures the overall acoustic intensity, whereas PNL better reflects the instantaneous perceived intensity of impulsive noise.

	\subsection{GRF predictor evaluation}
	We evaluate the GRF predictor on a held-out dataset. It contains four basic motion modes: (a) forward walking; (b) lateral walking; (c) backward walking; and (d) in-place turning. Other robot motions can be regarded as compositions of these four primitives. We report RMSE/MAE and the coefficient of determination $R^2$, and analyze performance across different motions. We conduct an ablation study by removing different loss terms to quantify their contributions:
	\begin{itemize}
		\item \textbf{C1 (Proposed method):} full loss function.
		\item \textbf{C2:} w/o $\mathcal{L}_{\mathrm{grf}}$.
		\item \textbf{C3:} w/o $\mathcal{L}_{\mathrm{dyn}}$.
		\item \textbf{C4:} w/o $\mathcal{L}_{\mathrm{swing}}$.
		\item \textbf{C5:} w/o $\mathcal{L}_{\mathrm{smooth}}$.
	\end{itemize}
	
	\begin{figure}[H]
		\centering
		\includegraphics[width=0.99\columnwidth]{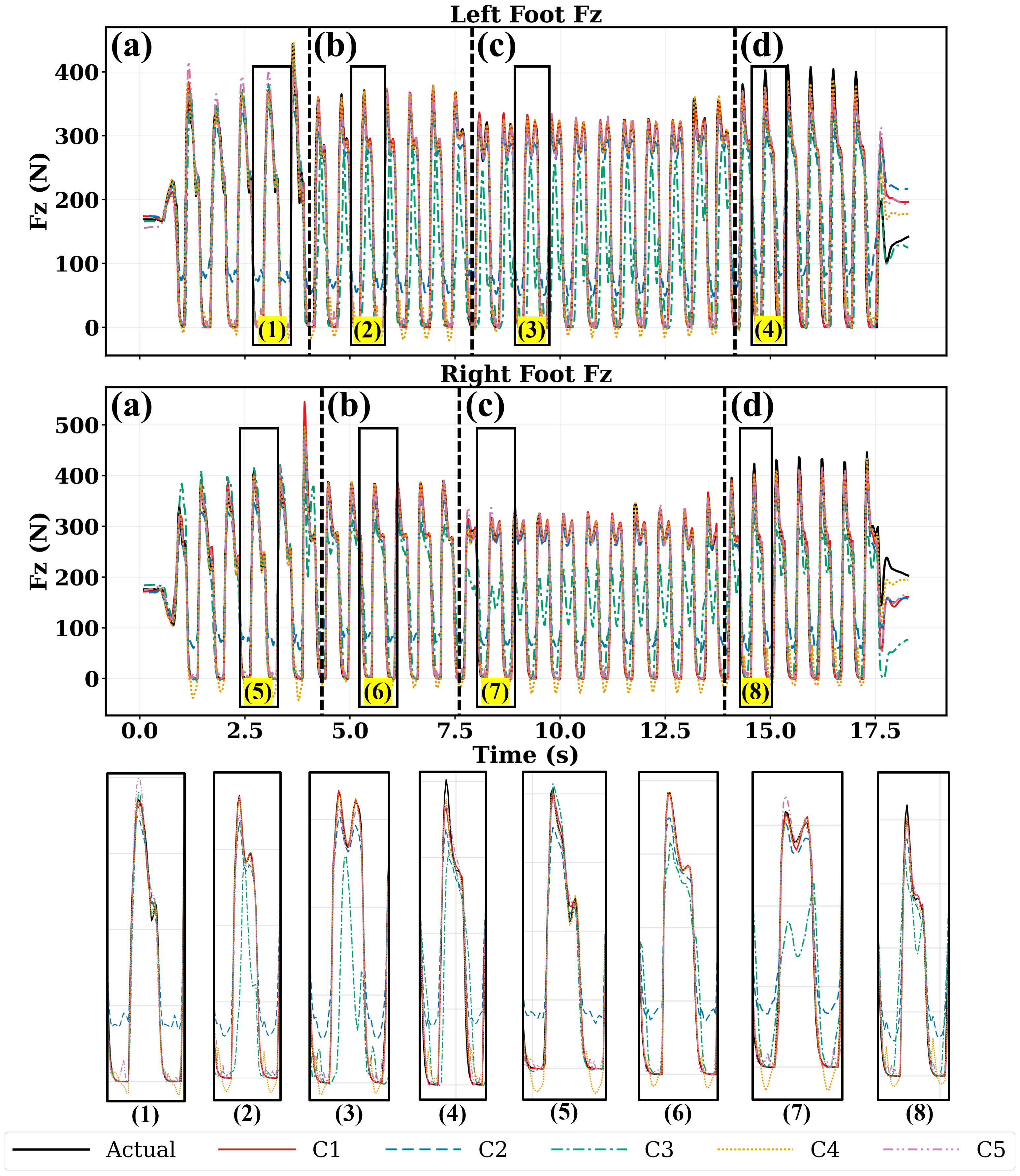}
		\caption{Ablation study of the GRF predictor under different loss configurations (C1–C5).}
		\label{fig:experiment-grf}
		\vspace{-0.8em} 
	\end{figure}
	
	As shown in Fig.~\ref{fig:experiment-grf}, C2 without the supervised term exhibits a significant drift in magnitude calibration. However, because the dynamics constraints remain, its overall temporal shape still matches the ground-truth curve well. In contrast, C3 without the dynamics residual clearly breaks physical self-consistency and reduces generalization across gaits. For C4, non-zero ``spurious forces'' appear during swing, leading to mis-detection of contact peaks and blurred stance–swing boundaries. Removing the smoothing regularizer in C5 introduces evident high-frequency oscillations around swing segments and contact transitions, degrading temporal stability.
	
	These qualitative observations are corroborated by the quantitative results in Table~\ref{tab:grf_ablation_walk_square}. Removing the inverse-dynamics residual ($\mathcal{L}_{\mathrm{dyn}}$, C3) causes the most severe degradation: the left/right RMSE increases from 14.49/14.00\,N to 106.29/79.44\,N (a $6$--$7\times$ increase), and $R^2$ drops from 0.9887/0.9899 to 0.3899/0.6747. In comparison, removing $\mathcal{L}_{\mathrm{swing}}$ (C4) or $\mathcal{L}_{\mathrm{smooth}}$ (C5) results in moderate error increases, consistent with the observed spurious swing-phase forces and temporal jitter.
	
	\begin{table}[t]
		\vspace*{3mm}
		\centering
		\caption{Ablation study of the GRF predictor on the held-out set.  \\Best results are highlighted in bold. \\PM means Proposed method. \\RMSE and MAE are reported in N.}
		\label{tab:grf_ablation_walk_square}
		\setlength{\tabcolsep}{3pt}
		\renewcommand{\arraystretch}{1.15}
		\begin{tabular}{l
				S[table-format=3.4] S[table-format=3.4] S[table-format=1.4]
				S[table-format=3.4] S[table-format=3.4] S[table-format=1.4]}
			\toprule
			\multirow{2}{*}{Config} & \multicolumn{3}{c}{Left foot} & \multicolumn{3}{c}{Right foot} \\
			\cmidrule(lr){2-4}\cmidrule(lr){5-7}
			& {RMSE$\downarrow$} & {MAE$\downarrow$} & {$R^2\uparrow$}
			& {RMSE$\downarrow$} & {MAE$\downarrow$} & {$R^2\uparrow$} \\
			\midrule
			C1(PM) &
			\bfseries 14.4861 & \bfseries  8.8441 & \bfseries 0.9887 &
			\bfseries 13.9972 & \bfseries  9.7272 & \bfseries 0.9899 \\
			C2 & 50.1127 & 40.7405 & 0.8644 & 49.3502 & 40.6566 & 0.8745 \\
			C3 & 106.2928 & 66.1469 & 0.3899 & 79.4400 & 58.0038 & 0.6747 \\
			C4 & 22.2752 & 10.5138 & 0.9732 & 20.2788 &  9.7324 & 0.9788 \\
			C5 & 26.2578 & 15.7987 & 0.9628 & 22.1852 & 12.9921 & 0.9746 \\
			\bottomrule
		\end{tabular}
	\end{table}

	\subsection{Quiet Locomotion}
	We evaluate locomotion noise and gait performance of different control strategies on the real robot at an average forward speed of 1.2\,m/s, and examine robustness under different footwear conditions. This speed corresponds to a moderately fast walking regime that elicits pronounced impact transients while remaining within the robot’s stable locomotion envelope. Moreover, it is representative of practical indoor service scenarios in human-centered environments, thereby providing a realistic, challenging, and safe test condition for acoustic evaluation.
	
	Experiments are conducted under four footwear conditions: barefoot, skate shoes, athletic sneakers, and high-heeled shoes, and four surface conditions: concrete, carpet, wood, and yoga mat. Note that, due to limitations in the recording chain and amplitude calibration, the reported A-weighted levels are used for relative comparison under a consistent recording setup. In addition, all results are obtained from five repeated trials, and we analyze the mean values. Detailed acoustic time-series curves are provided in the supplementary video due to space constraints.
	
	\begin{figure}[H]
		\centering
		\includegraphics[width=0.98\columnwidth]{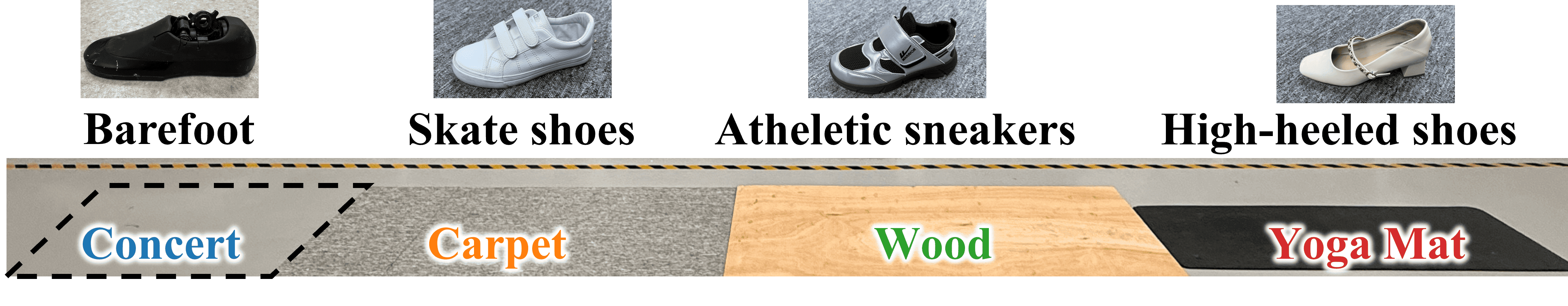}
		\caption{Footwear and surface conditions.}
		\label{fig:any-conditions}
		\vspace{-0.8em} 
	\end{figure}
	
	We compare the following three control strategies:
	\begin{itemize}
		\item \textbf{D1 (Baseline RL):} a standard RL walking policy without the quiet-walking curriculum.
		\item \textbf{D2 (Proposed Quiet RL):} the proposed quiet-walking policy.
		\item \textbf{D3 (Unitree Built-in):} the official built-in locomotion controller from Unitree, used as an engineering baseline.
	\end{itemize}
	
	\begin{figure}[H]
		\vspace*{3mm} 
		\centering
		\includegraphics[width=0.98\columnwidth]{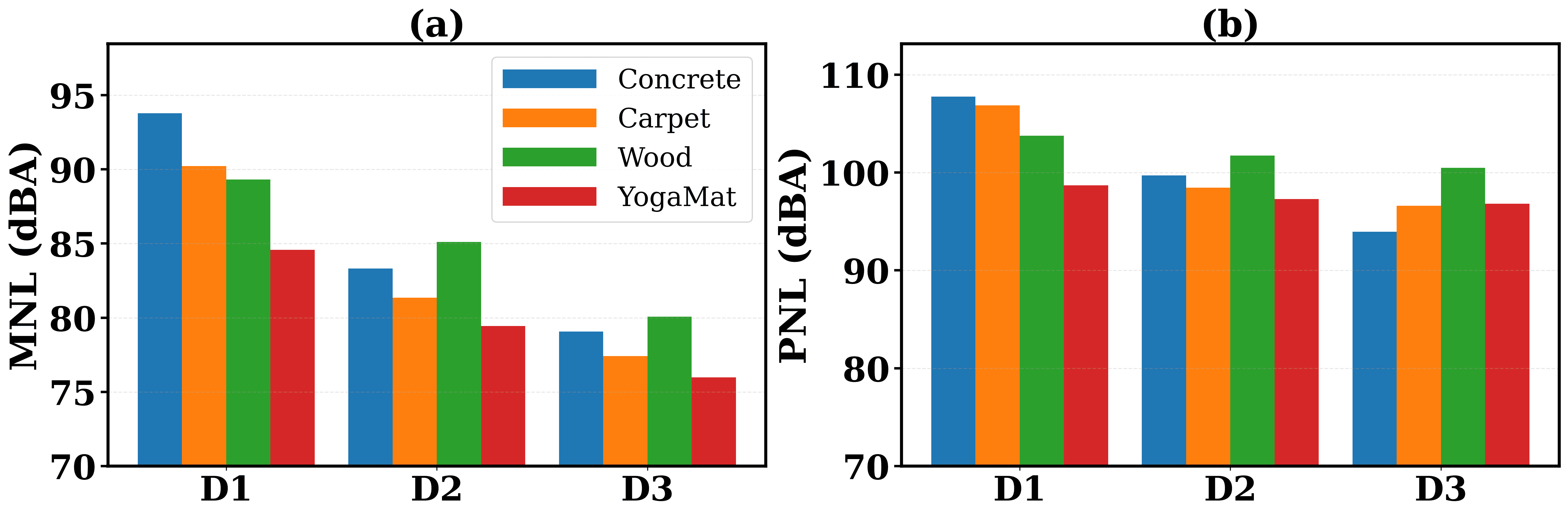}
		\caption{Mean noise level (MNL) and peak noise level (PNL) of D1, D2, and D3 under the barefoot condition across four surface conditions.}
		\label{fig:experiment-quiet}
		\vspace{-0.8em} 
	\end{figure}
	
	Taking barefoot as an example, Fig.~\ref{fig:experiment-quiet} shows that D2 significantly reduces noise compared with D1 across surface conditions. Averaged over the four surfaces, MNL decreases from 89.47\,dBA to 82.30\,dBA (a reduction of 7.17\,dB), and PNL decreases from 104.28\,dBA to 99.30\,dBA (a reduction of 4.98\,dB). The denoising gain is surface-dependent. Improvements are larger on concrete and carpet, and smaller on wood and yoga mat. This suggests that surface material is a dominant factor affecting noise. Moreover, D2 is overall close to the engineering baseline D3. Averaged over the four surfaces, the gaps between D2 and D3 are 4.17\,dB (MNL) and 2.34\,dB (PNL). In particular, the PNL gap on carpet/wood/yoga mat shrinks to 0.47--1.86\,dB. In addition, D3 exhibits smoother gait characteristics during contact transitions. Since its implementation is closed-source, we do not attribute its internal mechanisms, and use it only as an engineering baseline.
	
	\begin{figure}[H]
		\centering
		\includegraphics[width=0.98\columnwidth]{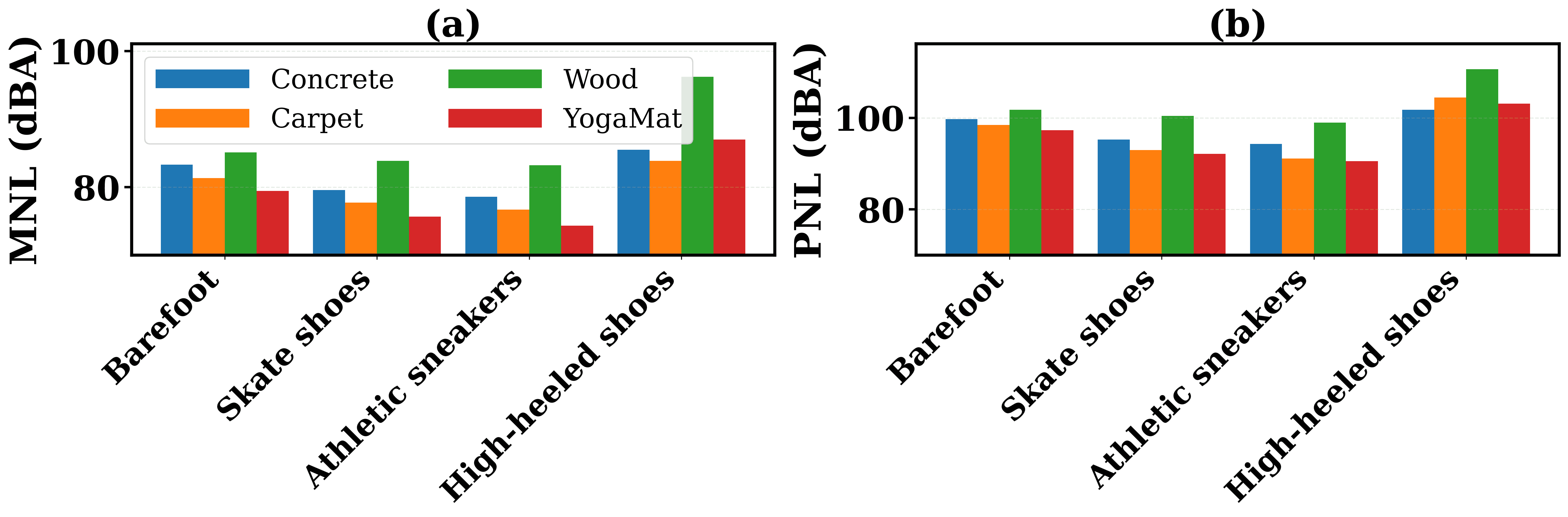}
		\caption{MNL and PNL of D2 across four footwear conditions and four surface conditions.}
		\label{fig:experiment-footwear}
		\vspace{-0.8em} 
	\end{figure}
	
	Under the same recording setup, Fig.~\ref{fig:experiment-footwear} reveals clear and interpretable structural patterns in the relative noise of D2 across footwear--surface combinations. Overall, footwear with stronger cushioning/damping and larger contact area (e.g., skate shoes and athletic sneakers) tends to yield lower MNL/PNL, whereas high-heeled shoes are noisier on most surfaces. Mechanistically, the shoe--ground system can be viewed as a coupling of effective contact stiffness and damping. Higher effective stiffness and shorter contact transitions concentrate impact energy at higher frequencies and more easily excite structural vibrations, which increases A-weighted levels and peak noise. In contrast, compliant and highly damped contact prolongs force rise time, suppresses high-frequency components, and reduces peaks. Surface materials also reflect differences in ``structural resonance/energy dissipation'' capability. Wood more easily amplifies impact-related components due to low damping, while the yoga mat substantially attenuates impact transmission via energy dissipation. Notably, the noise difference between the most adverse combination (high-heeled shoes $\times$ wood) and the most favorable combination (athletic sneakers $\times$ yoga mat) is close to 20\,dB. This indicates a strong interaction between footwear and surface. It also suggests that, under high-stiffness/low-damping contact conditions, the proposed quiet policy still faces a more challenging acoustic regime.
	
	\begin{figure}[H]
		\centering
		\includegraphics[width=0.8\columnwidth]{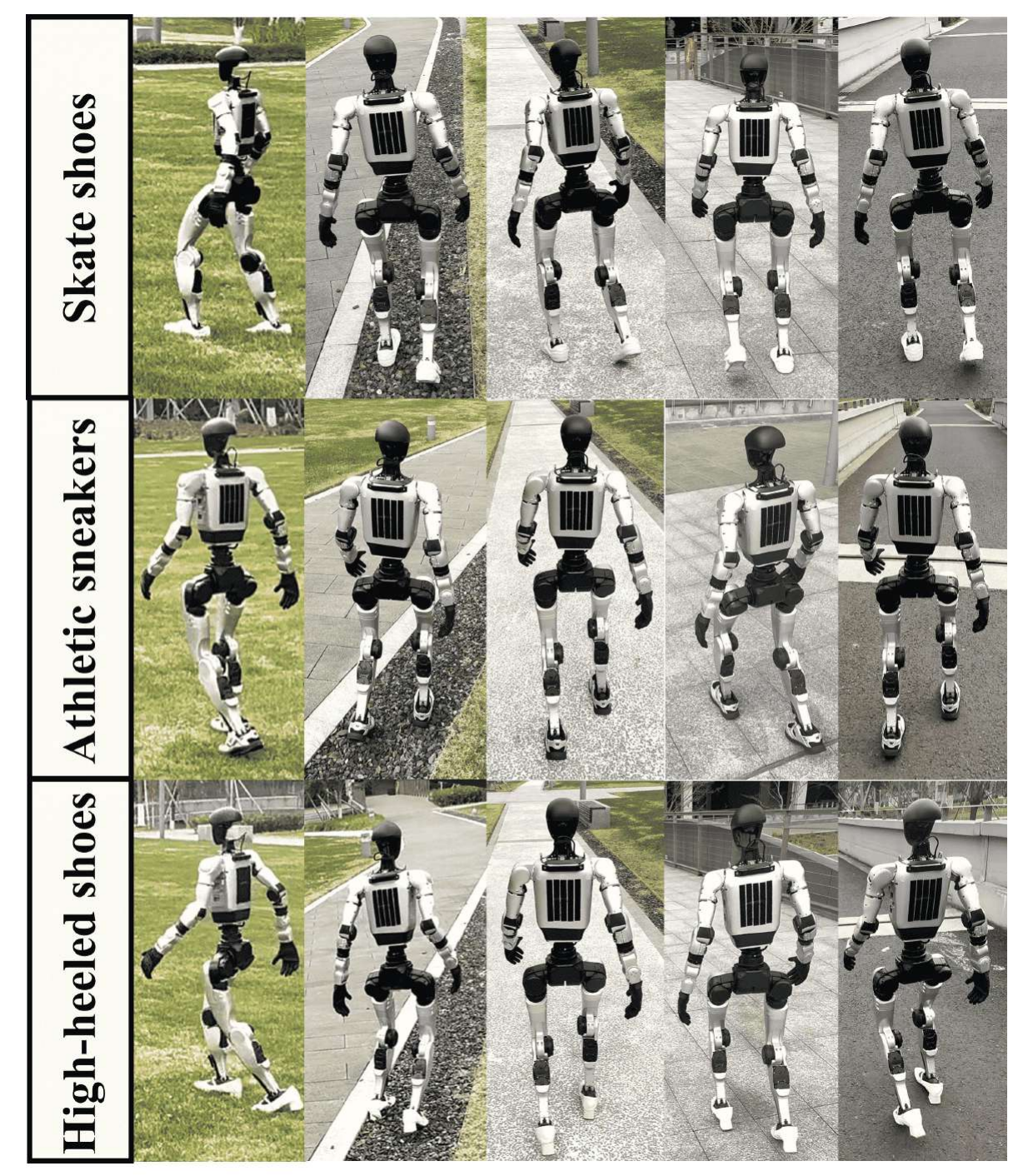}
		\caption{Outdoor robustness evaluation under diverse footwear conditions.}
		\label{fig:experiment-shoes}
		\vspace{-0.8em} 
	\end{figure}
	
	We further evaluate the robustness of D2 under diverse footwear conditions. As shown in Fig.~\ref{fig:experiment-shoes}, the robot performs walking tests on multiple outdoor terrains, including grass, gravel, cobblestone roads, smooth stone slab, and asphalt, under different footwear conditions. The policy maintains a stable gait without obvious posture divergence, repeated trips, or persistent slipping. This demonstrates good adaptability and transferability to variations in shoe--ground contact parameters.

	\section{CONCLUSION}
	This paper presents QuietWalk, a physics-informed reinforcement learning framework that enables deployable low-impact and low-noise humanoid locomotion across different footwear conditions. By coupling a PINN-based ground reaction force (GRF) estimator that satisfies inverse-dynamics consistency with policy optimization, the method embeds physically constrained force estimation directly into reward design, eliminating the need for external force sensors at deployment. This design combines model-driven structural constraints with data-driven learning, and provides impact-aware feedback signals while remaining practically deployable.
	
	Extensive ablation studies and comparative experiments show that dynamics-consistency constraints are critical for reliable force prediction and stable reward design. The resulting policy achieves stable acoustic-noise reduction across diverse surfaces and footwear conditions while maintaining robust locomotion performance. This indicates that the physics-informed impact regulation mechanism can generalize to non-nominal contact scenarios.
	
	The reported acoustic measurements are based on a fixed recording setup and are intended for consistent relative comparison under controlled conditions. In addition, the current framework mainly focuses on the vertical GRF component and does not explicitly model tangential contact interactions. Future work will extend force modeling and reward design to full 3D contact dynamics, improve acoustic-measurement calibration to enhance cross-platform comparability, and further explore deployable physics-informed locomotion control policies without specialized force sensors.

	\bibliographystyle{IEEEtran}
	\bibliography{references}

\end{document}